\documentclass[11pt,a4paper]{article}
\usepackage[hyperref]{acl2020}

\hypersetup{
 	pdfinfo={
 		Title={Pretrained Transformers Improve Out-of-Distribution Robustness},
 		Author={Dan Hendrycks and Xiaoyuan Liu and Eric Wallace and Adam Dziedzic and Rishabh Krishnan and Dawn Song},
 		Subject={NLP, Deep Learning, Robustness, Uncertainty, Adversarial Examples},
 		Keywords={BERT, robust, robustness, adversarial robustness, adversarial examples, transformers, OOD, out of distribution, distribution shift, anomaly detection, OOD detection, calibration}
 	}
}

\usepackage{times}
\usepackage{latexsym}

\usepackage{graphicx}
\usepackage{enumitem}
\usepackage{microtype}
\usepackage{subcaption}
\usepackage{textcomp}
\usepackage{booktabs}
\usepackage{gensymb}
\usepackage{mathtools}
\usepackage{amssymb}
\usepackage{wrapfig}
\usepackage{stfloats}
\usepackage{multirow}
\usepackage{tabularx}
\newcolumntype{Y}{>{\centering\arraybackslash}X}
\newcolumntype{L}{>{\raggedright\arraybackslash}X}

\usepackage{multirow}
\usepackage{arydshln}
\usepackage{makecell}
\usepackage{cleveref}

\aclfinalcopy 
\def\aclpaperid{2853} 

\newif\ifcomments
\ifcomments
    \providecommand{\eric}[1]{{\protect\color{magenta}{[Eric: #1]}}}
\else
    \providecommand{\eric}[1]{}
\fi

\title{Pretrained Transformers Improve Out-of-Distribution Robustness}

\author{\makecell{Dan Hendrycks$^1$\thanks{\; Equal contribution.\newline \url{https://github.com/camelop/NLP-Robustness}} \hspace{1cm} Xiaoyuan Liu$^{1,2}$\footnotemark[1] \hspace{1cm} Eric Wallace$^1$ \\ Adam Dziedzic$^3$ \hspace{0.82cm} Rishabh Krishnan$^1$ \hspace{0.8cm} Dawn Song$^1$} \\
$^1$UC Berkeley $^2$Shanghai Jiao Tong University $^3$University of Chicago\\
\{\href{mailto:hendrycks@berkeley.edu}{\tt hendrycks},\href{mailto:ericwallace@berkeley.edu}{\tt ericwallace},\href{mailto:dawnsong@berkeley.edu}{\tt dawnsong}\}{\tt @berkeley.edu}}
\date{}

\begin{document}
\maketitle
\begin{abstract}
Although pretrained Transformers such as BERT achieve high accuracy on in-distribution examples, do they generalize to new distributions? We systematically measure out-of-distribution (OOD) generalization for seven NLP datasets by constructing a new robustness benchmark with realistic distribution shifts. We measure the generalization of previous models including bag-of-words models, ConvNets, and LSTMs, and we show that pretrained Transformers' performance declines are substantially smaller. Pretrained transformers are also more effective at detecting anomalous or OOD examples, while many previous models are frequently \textit{worse} than chance. We examine which factors affect robustness, finding that larger models are not necessarily more robust, distillation can be harmful, and more diverse pretraining data can enhance robustness. Finally, we show where future work can improve OOD robustness.\looseness=-1
\end{abstract}

\section{Introduction}

The train and test distributions are often not identically distributed. Such train-test mismatches occur because evaluation datasets rarely characterize the entire distribution~\cite{Torralba2011UnbiasedLA}, and the test distribution typically drifts over time~\cite{QuioneroCandela2009DatasetSI}. Chasing an evolving data distribution is costly, and even if the training data does not become stale, models will still encounter unexpected situations at test time.
Accordingly, 
models must \emph{generalize} to OOD examples whenever possible, and when OOD examples do not belong to any known class, models must \emph{detect} them in order to abstain or trigger a conservative fallback policy~\cite{Emmott2015AMO}.

Most evaluation in natural language processing (NLP) assumes the train and test examples are independent and identically distributed (IID). In the IID setting, large pretrained Transformer models can attain near human-level performance on numerous tasks \cite{Wang2018GLUEA}. However, high IID accuracy does not necessarily translate to OOD robustness for image classifiers~\cite{Hendrycks2019BenchmarkingNN}, and pretrained Transformers may embody this same fragility. 
Moreover,
pretrained Transformers can rely heavily on spurious cues and annotation artifacts \cite{Cai2017PayAT,gururangan2018annotation} which out-of-distribution examples are less likely to include, so their OOD robustness remains uncertain.

In this work, we systematically study the OOD robustness of various NLP models, such as word embeddings averages, LSTMs, pretrained Transformers, and more. We decompose OOD robustness into a model's ability to (1) generalize and to (2) detect OOD examples~\cite{Card2018DeepWA}.

To measure OOD generalization, we create a new evaluation benchmark that tests robustness to shifts in writing style, topic, and vocabulary, and spans the tasks of sentiment analysis, textual entailment, question answering, and semantic similarity. We create OOD test sets by splitting datasets with their metadata or by pairing similar datasets together (Section~\ref{sec:setup}). Using our OOD generalization benchmark, we show that pretrained Transformers are considerably more robust to OOD examples than traditional NLP models (Section~\ref{sec:accuracy}). We show that the performance of an LSTM semantic similarity model declines by over $35\%$ on OOD examples, while a RoBERTa model's performance slightly \textit{increases}. Moreover, we demonstrate that while pretraining larger models does not seem to improve OOD generalization, pretraining models on diverse data does improve OOD generalization.

To measure OOD detection performance, we turn classifiers into anomaly detectors by using their prediction confidences as anomaly scores \cite{Hendrycks2016ABF}. We show that many non-pretrained NLP models are often near or \emph{worse than random chance} at OOD detection. In contrast, pretrained Transformers are far more capable at OOD detection. Overall, our results highlight that while there is room for future robustness improvements, pretrained Transformers are already moderately robust.\looseness=-1
\section{How We Test Robustness}\label{sec:setup}

\subsection{Train and Test Datasets}

We evaluate OOD generalization with \emph{seven} carefully selected datasets. 
Each dataset either (1) contains metadata which allows us to naturally split the samples or (2) can be paired with a similar dataset from a distinct data generating process. By splitting or grouping our chosen datasets, we can induce a distribution shift and measure OOD generalization.

\noindent We utilize four sentiment analysis datasets:
\begin{itemize}[nosep,leftmargin=4mm]
    \item We use \textbf{SST-2}, which contains pithy expert movie reviews~\cite{socher2013recursive}, and \textbf{IMDb}~\cite{maas2011learning}, which contains full-length lay movie reviews. We train on one dataset and evaluate on the other dataset, and vice versa. Models predict a movie review's binary sentiment, and we report accuracy.
    \item The \textbf{Yelp Review Dataset} contains restaurant reviews with detailed metadata (e.g., user ID, restaurant name).
    We carve out four groups from the dataset based on food type: \emph{American, Chinese, Italian,} and \emph{Japanese}. Models predict a restaurant review's binary sentiment, and we report accuracy.
    \item The \textbf{Amazon Review Dataset} contains product reviews from Amazon~\cite{DBLP:journals/corr/McAuleyTSH15,DBLP:journals/corr/HeM16}. We split the data into five categories of clothing (Clothes, Women Clothing, Men Clothing, Baby Clothing, Shoes) and two categories of entertainment products (Music, Movies). We sample 50,000 reviews for each category. Models predict a review's 1 to 5 star rating, and we report accuracy.
\end{itemize}

\noindent We also utilize these datasets for semantic similarity, reading comprehension, and textual entailment:
\begin{itemize}[nosep,leftmargin=4mm]
    \item \textbf{STS-B} requires predicting the semantic similarity between pairs of sentences~\cite{cer2017semeval}. The dataset contains text of different genres and sources; we use four  sources from two genres: MSRpar (news), Headlines (news); MSRvid (captions), Images (captions). The evaluation metric is Pearson's correlation coefficient.
    \item \textbf{ReCoRD} is a reading comprehension dataset using paragraphs from CNN and Daily Mail news articles and automatically generated questions~\cite{DBLP:journals/corr/abs-1810-12885}. We bifurcate the dataset into CNN and Daily Mail splits and evaluate using exact match.
    \item \textbf{MNLI} is a textual entailment dataset using sentence pairs drawn from different genres of text~\cite{williams2017broad}. We select examples from two genres of transcribed text (Telephone and Face-to-Face) and one genre of written text (Letters), and we report classification accuracy.
\end{itemize}

\subsection{Embedding and Model Types}

We evaluate NLP models with different input representations and encoders. We investigate three model categories with a total of thirteen models.

\paragraph{Bag-of-words (BoW) Model.} We use a bag-of-words model~\cite{harris1954distributional}, which is high-bias but low-variance, so it may exhibit performance stability. The BoW model is only used for sentiment analysis and STS-B due to its low performance on the other tasks. For STS-B, we use the cosine similarity of the BoW representations from the two input sentences.

\paragraph{Word Embedding Models.} We use word2vec~\cite{mikolov2013distributed} and GloVe~\cite{pennington2014glove} word embeddings. These embeddings are encoded with one of three models: word averages~\cite{Wieting2015TowardsUP}, LSTMs~\cite{hochreiter1997lstm}, and Convolutional Neural Networks (ConvNets). For classification tasks, the representation from the encoder is fed into an MLP. For STS-B and MNLI, we use the cosine similarity of the encoded representations from the two input sentences. For reading comprehension, we use the DocQA model~\cite{clark2017simple} with GloVe embeddings. We implement our models in AllenNLP~\cite{Gardner2017AllenNLP} and tune the hyperparameters to maximize validation performance on the IID task.

\paragraph{Pretrained Transformers.} We investigate BERT-based models~\cite{Devlin2019BERTPO} which are pretrained bidirectional Transformers~\cite{vaswani2017attention} with GELU~\cite{hendrycks2016gelu} activations. In addition to using BERT Base and BERT Large, we also use the large version of RoBERTa~\cite{liu2019roberta}, which is pretrained on a larger dataset than BERT. We use ALBERT~\cite{Lan2020ALBERT} and also a distilled version of BERT, DistilBERT~\cite{sanh2019distilbert}. We follow the standard BERT fine-tuning procedure~\cite{Devlin2019BERTPO} and lightly tune the hyperparameters for our tasks. We perform our experiments using the HuggingFace Transformers library~\cite{Wolf2019HuggingFacesTS}.
\section{Out-of-Distribution Generalization}\label{sec:accuracy}

\begin{figure}[t]
\centering
\vspace{-0.35cm}
\includegraphics[width=0.5\textwidth]{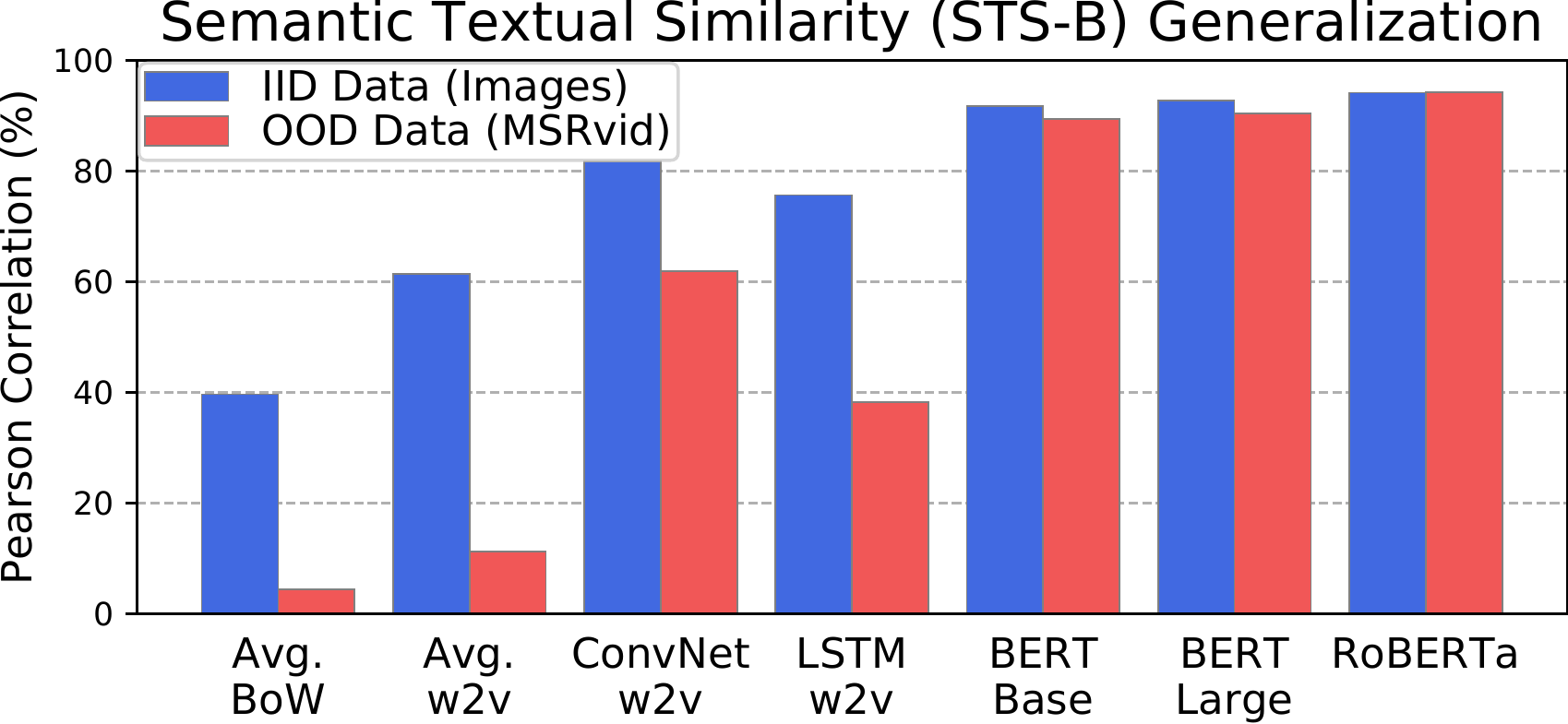}
\vspace{-15pt}
\caption{Pretrained Transformers often have smaller IID/OOD generalization gaps than previous models.}
\label{fig:sts}
\vspace{-10pt}
\end{figure}

In this section, we evaluate OOD generalization of numerous NLP models on seven datasets and provide some upshots. A subset of results are in Figures~\ref{fig:sts} and \ref{fig:ood_accuracy}. Full results are in Appendix~\ref{appendix:additional}.

\paragraph{Pretrained Transformers are More Robust.} In our experiments, pretrained Transformers often have smaller generalization gaps from IID data to OOD data than traditional NLP models. For instance, Figure~\ref{fig:sts} shows that the LSTM model declined by over 35\%, while RoBERTa's generalization performance in fact increases. For Amazon, MNLI, and Yelp, we find that pretrained Transformers' accuracy only slightly fluctuates on OOD examples. Partial MNLI results are in Table~\ref{tab:multinli}. We present the full results for these three tasks in Appendix~\ref{appendix:minor}. In short, pretrained Transformers can generalize across a variety of distribution shifts.\looseness=-1

\begin{table}[h]
\begin{center}
\begin{tabular}{l|ccc}
Model & \multicolumn{1}{p{1.5cm}}{\centering Telephone (IID)} & \multicolumn{1}{p{1.5cm}}{\centering Letters (OOD)} & \multicolumn{1}{p{2cm}}{\centering Face-to-Face (OOD)}\\
\hline
BERT & 81.4\% & 82.3\% & 80.8\%\\
\bottomrule
\end{tabular}
\end{center}
\caption{Accuracy of a BERT Base MNLI model trained on Telephone data and tested on three different distributions. Accuracy only slightly fluctuates.}
\label{tab:multinli}
\vspace{-5pt}
\end{table}

\begin{figure}[t]
\centering
\begin{subfigure}{0.5\textwidth}
\vspace{-0.35cm}
\includegraphics[width=\textwidth]{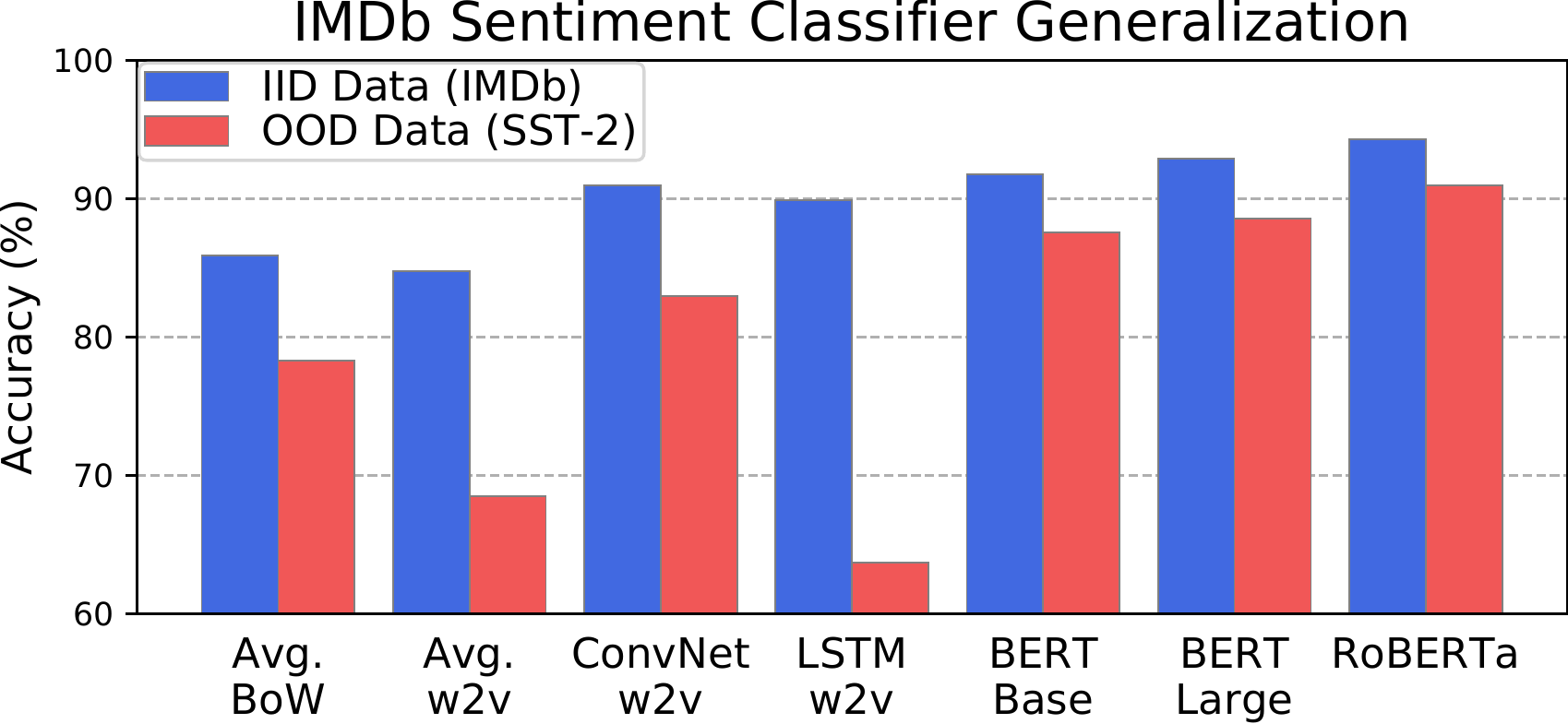}
\label{fig:imdb}
\end{subfigure}
\begin{subfigure}{0.5\textwidth}
\vspace{-0.15cm}
\includegraphics[width=\textwidth]{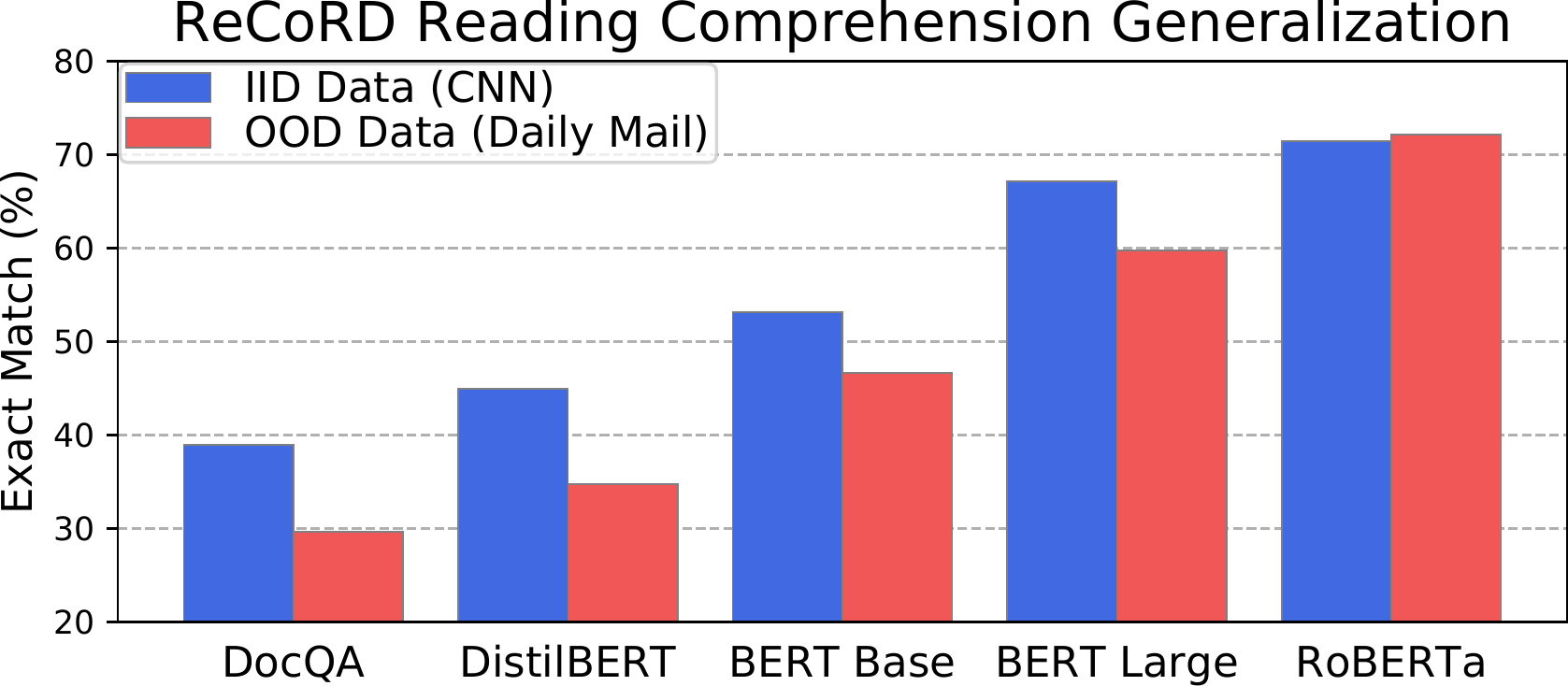}
\label{fig:record}
\end{subfigure}
\vspace{-15pt}
\caption{Generalization results for sentiment analysis and reading comprehension. While IID accuracy does not vary much for IMDb sentiment analysis, OOD accuracy does. Here pretrained Transformers do best.}
\label{fig:ood_accuracy}
\end{figure}

\begin{figure}[h!]
\centering
\includegraphics[width=0.48\textwidth]{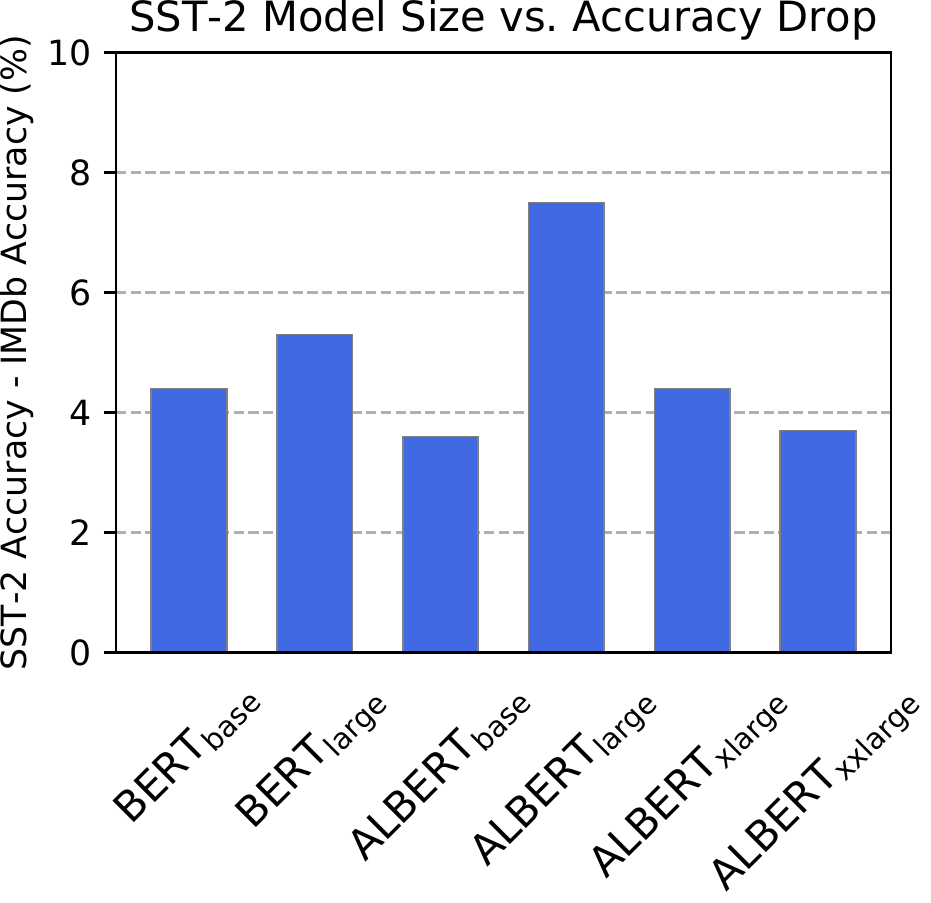}
\vspace{-15pt}
\caption{The IID/OOD generalization gap is not improved with larger models, unlike in computer vision.}
\label{fig:size}
\vspace{-5pt}
\end{figure}

\paragraph{Bigger Models Are Not Always Better.} While larger models reduce the IID/OOD generalization gap in computer vision \cite{Hendrycks2019BenchmarkingNN,Xie2019IntriguingPO,Hendrycks2019NaturalAE}, we find the same does \emph{not} hold in NLP. Figure~\ref{fig:size} shows that larger BERT and ALBERT models do not reduce the generalization gap.
However, in keeping with results from vision \cite{Hendrycks2019BenchmarkingNN}, we find that model distillation can reduce robustness, as evident in our DistilBERT results in Figure~\ref{fig:ood_accuracy}. This highlights that testing model compression methods for BERT~\cite{shen2019q,ganesh2020compressing,li2020train}
on only in-distribution examples gives a limited account of model generalization, and such narrow evaluation may mask downstream costs.

\paragraph{More Diverse Data Improves Generalization.} Similar to computer vision~\cite{Orhan2019RobustnessPO, Xie2019SelftrainingWN,hendrycks2019pretraining}, pretraining on larger and more diverse datasets can improve robustness. RoBERTa exhibits greater robustness than BERT Large, where one of the largest differences between these two models is that RoBERTa pretrains on more data. See Figure~\ref{fig:ood_accuracy}'s results.

\begin{figure*}
    \vspace{-20pt}
    \centering
    \includegraphics[width=\textwidth]{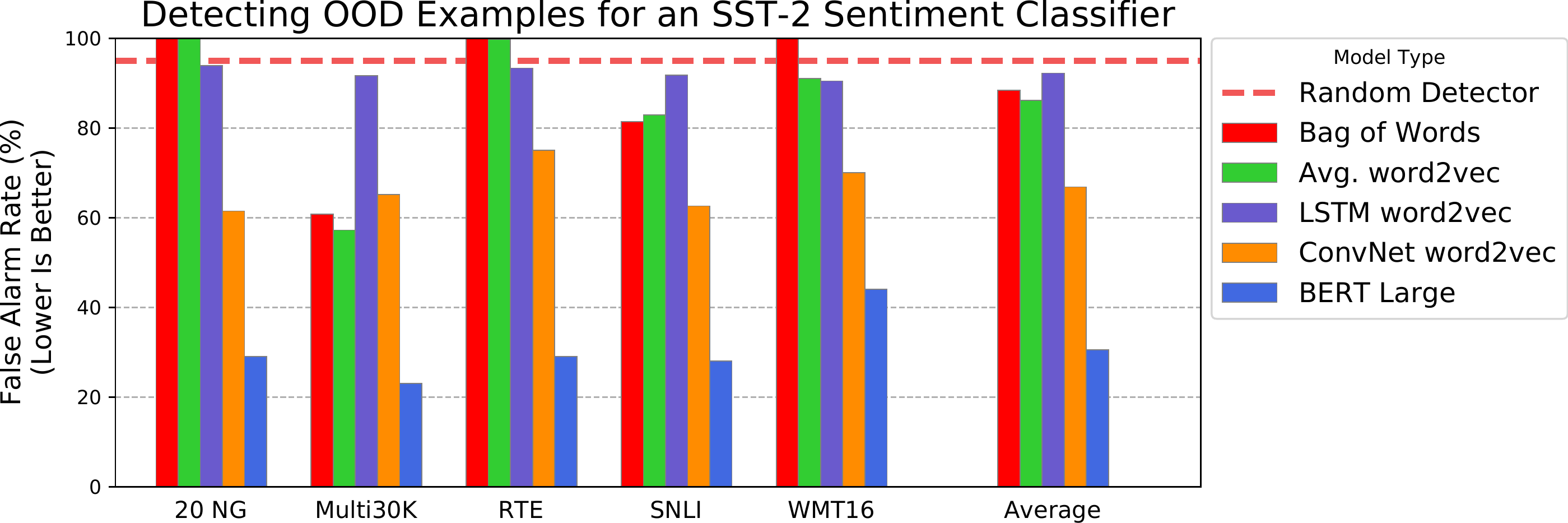}
    \caption{We feed in OOD examples from out-of-distribution datasets (20 Newsgroups, Multi30K, etc.) to SST-2 sentiment classifiers and report the False Alarm Rate at $95\%$ Recall. A lower False Alarm Rate is better. Classifiers are repurposed as anomaly detectors by using their negative maximum softmax probability as the anomaly score---OOD examples should be predicted with less confidence than IID examples. Models such as BoW, word2vec averages, and LSTMs are near random chance; that is, previous NLP models are frequently more confident when classifying OOD examples than when classifying IID test examples.
    }
    \label{fig:fpr}
    \vspace{-10pt}
\end{figure*}

\section{Out-of-Distribution Detection}\label{sec:ood_detection}

Since OOD robustness requires evaluating both OOD generalization and OOD detection, we now turn to the latter. Without access to an outlier dataset~\cite{Hendrycks2018DeepAD}, the state-of-the-art OOD detection technique is to use the model's prediction confidence to separate in- and out-of-distribution examples \cite{Hendrycks2016ABF}. Specifically, we assign an example $x$ the anomaly score $-\max_y p(y\mid x)$, the negative prediction confidence, to perform OOD detection.

We train models on SST-2, record the model's confidence values on SST-2 test examples, and then record the model's confidence values on OOD examples from five other datasets. For our OOD examples, we use validation examples from 20 Newsgroups (20 NG)~\cite{Lang95}, the English source side of English-German WMT16 and English-German Multi30K~\cite{elliott2016multi30k}, and concatenations of the premise and hypothesis for RTE~\cite{dagan2005pascal} and SNLI~\cite{Bowman2015ALA}.
These examples are only used during OOD evaluation not training. 

For evaluation, we follow past work~\cite{Hendrycks2018DeepAD} and report the False Alarm Rate at $95\%$ Recall (FAR95). The FAR95 is the probability that an in-distribution example raises a false alarm, assuming that 95\% of all out-of-distribution examples are detected. Hence a lower FAR95 is better.
Partial results are in Figure~\ref{fig:fpr}, and full results are in Appendix~\ref{appendix:ooddetection}.

\paragraph{Previous Models Struggle at OOD Detection.} Models without pretraining (e.g., BoW, LSTM word2vec) are often unable to reliably detect OOD examples. In particular, these models' FAR95 scores are sometimes \emph{worse} than chance because the models often assign a higher probability to out-of-distribution examples than in-distribution examples. The models particularly struggle on 20 Newsgroups (which contains text on diverse topics including computer hardware, motorcycles, space), as their false alarm rates are approximately $100\%$.

\begin{figure}
    \centering
    \includegraphics[width=0.48\textwidth]{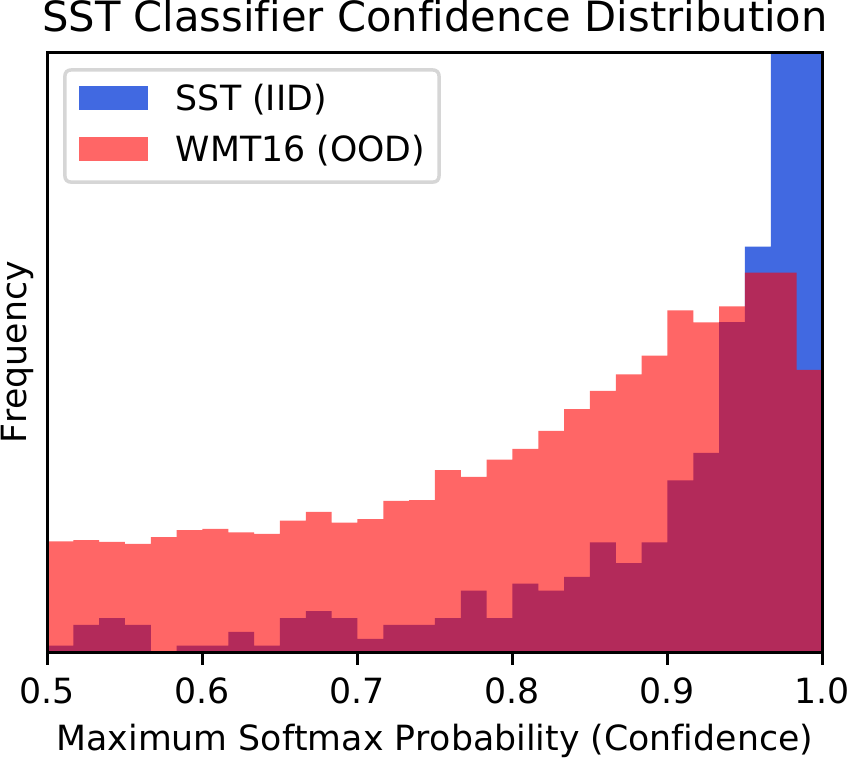}
    \caption{The confidence distribution for a RoBERTa SST-2 classifier on examples from the SST-2 test set and the English side of WMT16 English-German. The WMT16 histogram is translucent and overlays the SST histogram. The minimum prediction confidence is $0.5$. Although RoBERTa is better than previous models at OOD detection, there is clearly room for future work.}
    \label{fig:roberta_ood}
\end{figure}

\paragraph{Pretrained Transformers Are Better Detectors.} In contrast, pretrained Transformer models are better OOD detectors. Their FAR95 scores are always better than chance.
Their superior detection performance is not solely because the underlying model is a language model, as prior work~\cite{Hendrycks2018DeepAD}
shows that language models are not necessarily adept at OOD detection. Also note that in OOD detection for computer vision, higher accuracy does not reliably improve OOD detection~\cite{kimin}, so pretrained Transformers' OOD detection performance is not anticipated.
Despite their relatively low FAR95 scores, pretrained Transformers still do not cleanly separate in- and out-of-distribution examples (Figure~\ref{fig:roberta_ood}). OOD detection using pretrained Transformers is still far from perfect, and future work can aim towards creating better methods for OOD detection.

\section{Discussion and Related Work}

\paragraph{Why Are Pretrained Models More Robust?} An interesting area for future work is to analyze \textit{why} pretrained Transformers are more robust. A flawed explanation is that pretrained models are simply more accurate. However, this work and past work show that increases in accuracy do not directly translate to reduced IID/OOD generalization gaps~\cite{Hendrycks2019BenchmarkingNN,fried2019cross}.
One partial explanation is that Transformer models are pretrained on \textit{diverse} data, and in computer vision, dataset diversity can improve OOD generalization~\cite{hendrycks2020augmix} and OOD detection~\cite{Hendrycks2018DeepAD}.
Similarly, Transformer models are pretrained with large \textit{amounts} of data, which may also aid robustness~\cite{Orhan2019RobustnessPO, Xie2019SelftrainingWN,hendrycks2019pretraining}. However, this is not a complete explanation as BERT is pretrained on roughly 3 billion tokens, while GloVe is trained on roughly 840 billion tokens. Another partial explanation may lie in self-supervised training itself. \citet{hendrycks2019self} show that computer vision models trained with self-supervised objectives exhibit better OOD generalization and far better OOD detection performance. Future work could propose new self-supervised objectives that enhance model robustness.\looseness=-1

\paragraph{Domain Adaptation.} Other research on robustness considers the separate problem of domain adaptation~\cite{Blitzer2007BiographiesBB,daume2009frustratingly}, where models must learn representations of a source and target distribution. We focus on testing generalization \textit{without} adaptation in order to benchmark robustness to unforeseen distribution shifts. Unlike \citet{fisch2019proceedings,Yogatama2019LearningAE}, we measure OOD generalization by considering simple and natural distribution shifts, and we also evaluate more than question answering.

\paragraph{Adversarial Examples.} Adversarial examples can be created for NLP models by inserting phrases~\cite{Jia2017AdversarialEF,Wallace2019UniversalAT}, paraphrasing questions~\cite{ribeiro2018semantically}, and reducing inputs~\cite{feng2018pathologies}. However, adversarial examples are often disconnected from real-world performance concerns~\cite{Gilmer2018MotivatingTR}. Thus, we focus on an experimental setting that is more realistic. While previous works show that, for all NLP models, there exist adversarial examples, we show that all models are not equally fragile. Rather, pretrained Transformers are overall far more robust than previous models.

\paragraph{Counteracting Annotation Artifacts.} Annotators can accidentally leave unintended shortcuts in datasets that allow models to achieve high accuracy by effectively ``cheating''~\cite{Cai2017PayAT,gururangan2018annotation,min2019compositional}. These \textit{annotation artifacts} are one reason for OOD brittleness: OOD examples are unlikely to contain the same spurious patterns as in-distribution examples. OOD robustness benchmarks like ours can \textit{stress test} a model's dependence on artifacts~\cite{liu2019inoculation,feng2019misleading,naik2018stress}.

\section{Conclusion}
We created an expansive benchmark across several NLP tasks to evaluate out-of-distribution robustness. To accomplish this, we carefully restructured and matched previous datasets to induce numerous realistic distribution shifts. We first showed that pretrained Transformers \emph{generalize} to OOD examples far better than previous models, so that the IID/OOD generalization gap is often markedly reduced.
We then showed that pretrained Transformers \emph{detect} OOD examples surprisingly well. Overall, our extensive evaluation shows that while pretrained Transformers are moderately robust, there remains room for future research on robustness.\looseness=-1
\newpage
\section*{Acknowledgements}

We thank the members of Berkeley NLP, Sona Jeswani, Suchin Gururangan, Nelson Liu, Shi Feng, the anonymous reviewers, and especially Jon Cai. This material is in part based upon work supported by the National Science Foundation Frontier Award 1804794. Any opinions, findings, and conclusions or recommendations expressed in this material are those of the author(s) and do not necessarily reflect the views of the National Science Foundation.

\bibliography{acl2020}

\begin{thebibliography}{60}
\expandafter\ifx\csname natexlab\endcsname\relax\def\natexlab#1{#1}\fi

\bibitem[{Blitzer et~al.(2007)Blitzer, Dredze, and
  Pereira}]{Blitzer2007BiographiesBB}
John Blitzer, Mark Dredze, and Fernando Pereira. 2007.
\newblock Biographies, bollywood, boom-boxes and blenders: Domain adaptation
  for sentiment classification.
\newblock In \emph{ACL}.

\bibitem[{Bowman et~al.(2015)Bowman, Angeli, Potts, and
  Manning}]{Bowman2015ALA}
Samuel~R. Bowman, Gabor Angeli, Christopher Potts, and Christopher~D. Manning.
  2015.
\newblock A large annotated corpus for learning natural language inference.
\newblock In \emph{EMNLP}.

\bibitem[{Cai et~al.(2017)Cai, Tu, and Gimpel}]{Cai2017PayAT}
Zheng Cai, Lifu Tu, and Kevin Gimpel. 2017.
\newblock Pay attention to the ending: Strong neural baselines for the roc
  story cloze task.
\newblock In \emph{ACL}.

\bibitem[{Card et~al.(2018)Card, Zhang, and Smith}]{Card2018DeepWA}
Dallas Card, Michael Zhang, and Noah~A. Smith. 2018.
\newblock Deep weighted averaging classifiers.
\newblock In \emph{FAT}.

\bibitem[{Cer et~al.(2017)Cer, Diab, Agirre, Lopez-Gazpio, and
  Specia}]{cer2017semeval}
Daniel Cer, Mona Diab, Eneko Agirre, Inigo Lopez-Gazpio, and Lucia Specia.
  2017.
\newblock {SemEval-2017} {Task} 1: Semantic textual similarity-multilingual and
  cross-lingual focused evaluation.
\newblock In \emph{SemEval}.

\bibitem[{Clark and Gardner(2018)}]{clark2017simple}
Christopher Clark and Matt Gardner. 2018.
\newblock Simple and effective multi-paragraph reading comprehension.
\newblock In \emph{ACL}.

\bibitem[{Dagan et~al.(2005)Dagan, Glickman, and Magnini}]{dagan2005pascal}
Ido Dagan, Oren Glickman, and Bernardo Magnini. 2005.
\newblock The {PASCAL} recognising textual entailment challenge.
\newblock In \emph{Machine Learning Challenges Workshop}.

\bibitem[{Daum{\'e}~III(2007)}]{daume2009frustratingly}
Hal Daum{\'e}~III. 2007.
\newblock Frustratingly easy domain adaptation.
\newblock In \emph{ACL}.

\bibitem[{Devlin et~al.(2019)Devlin, Chang, Lee, and
  Toutanova}]{Devlin2019BERTPO}
Jacob Devlin, Ming-Wei Chang, Kenton Lee, and Kristina Toutanova. 2019.
\newblock {BERT:} pre-training of deep bidirectional transformers for language
  understanding.
\newblock In \emph{NAACL-HLT}.

\bibitem[{Elliott et~al.(2016)Elliott, Frank, Sima'an, and
  Specia}]{elliott2016multi30k}
Desmond Elliott, Stella Frank, Khalil Sima'an, and Lucia Specia. 2016.
\newblock Multi30k: Multilingual english-german image descriptions.
\newblock In \emph{ACL}.

\bibitem[{Emmott et~al.(2015)Emmott, Das, Dietterich, Fern, and
  Wong}]{Emmott2015AMO}
Andrew Emmott, Shubhomoy Das, Thomas~G. Dietterich, Alan Fern, and Weng-Keen
  Wong. 2015.
\newblock A meta-analysis of the anomaly detection problem.

\bibitem[{Feng et~al.(2019)Feng, Wallace, and Boyd-Graber}]{feng2019misleading}
Shi Feng, Eric Wallace, and Jordan Boyd-Graber. 2019.
\newblock Misleading failures of partial-input baselines.
\newblock In \emph{ACL}.

\bibitem[{Feng et~al.(2018)Feng, Wallace, Grissom, Iyyer, Rodriguez, and
  Boyd-Graber}]{feng2018pathologies}
Shi Feng, Eric Wallace, II~Grissom, Mohit Iyyer, Pedro Rodriguez, and Jordan
  Boyd-Graber. 2018.
\newblock Pathologies of neural models make interpretations difficult.
\newblock In \emph{EMNLP}.

\bibitem[{Fisch et~al.(2019)Fisch, Talmor, Jia, Seo, Choi, and
  Chen}]{fisch2019proceedings}
Adam Fisch, Alon Talmor, Robin Jia, Minjoon Seo, Eunsol Choi, and Danqi Chen.
  2019.
\newblock Proceedings of the 2nd workshop on machine reading for question
  answering.
\newblock In \emph{MRQA Workshop}.

\bibitem[{Fried et~al.(2019)Fried, Kitaev, and Klein}]{fried2019cross}
Daniel Fried, Nikita Kitaev, and Dan Klein. 2019.
\newblock Cross-domain generalization of neural constituency parsers.
\newblock In \emph{ACL}.

\bibitem[{Ganesh et~al.(2020)Ganesh, Chen, Lou, Khan, Yang, Chen, Winslett,
  Sajjad, and Nakov}]{ganesh2020compressing}
Prakhar Ganesh, Yao Chen, Xin Lou, Mohammad~Ali Khan, Yin Yang, Deming Chen,
  Marianne Winslett, Hassan Sajjad, and Preslav Nakov. 2020.
\newblock Compressing large-scale transformer-based models: A case study on
  {BERT}.
\newblock \emph{ArXiv}, abs/2002.11985.

\bibitem[{Gardner et~al.(2018)Gardner, Grus, Neumann, Tafjord, Dasigi, Liu,
  Peters, Schmitz, and Zettlemoyer}]{Gardner2017AllenNLP}
Matt Gardner, Joel Grus, Mark Neumann, Oyvind Tafjord, Pradeep Dasigi,
  Nelson~F. Liu, Matthew Peters, Michael Schmitz, and Luke~S. Zettlemoyer.
  2018.
\newblock {AllenNLP:} a deep semantic natural language processing platform.
\newblock In \emph{Workshop for NLP Open Source Software}.

\bibitem[{Gilmer et~al.(2018)Gilmer, Adams, Goodfellow, Andersen, and
  Dahl}]{Gilmer2018MotivatingTR}
Justin Gilmer, Ryan~P. Adams, Ian~J. Goodfellow, David Andersen, and George~E.
  Dahl. 2018.
\newblock Motivating the rules of the game for adversarial example research.
\newblock \emph{ArXiv}, abs/1807.06732.

\bibitem[{Gururangan et~al.(2018)Gururangan, Swayamdipta, Levy, Schwartz,
  R.~Bowman, and A.~Smith}]{gururangan2018annotation}
Suchin Gururangan, Swabha Swayamdipta, Omer Levy, Roy Schwartz, Samuel
  R.~Bowman, and Noah A.~Smith. 2018.
\newblock Annotation artifacts in natural language inference data.
\newblock In \emph{NAACL-HLT}.

\bibitem[{Harris(1954)}]{harris1954distributional}
Zellig~S Harris. 1954.
\newblock Distributional structure.
\newblock \emph{Word}.

\bibitem[{He and McAuley(2016)}]{DBLP:journals/corr/HeM16}
Ruining He and Julian~J. McAuley. 2016.
\newblock Ups and downs: Modeling the visual evolution of fashion trends with
  one-class collaborative filtering.
\newblock In \emph{WWW}.

\bibitem[{Hendrycks and Dietterich(2019)}]{Hendrycks2019BenchmarkingNN}
Dan Hendrycks and Thomas Dietterich. 2019.
\newblock Benchmarking neural network robustness to common corruptions and
  perturbations.
\newblock In \emph{ICLR}.

\bibitem[{Hendrycks and Gimpel(2016)}]{hendrycks2016gelu}
Dan Hendrycks and Kevin Gimpel. 2016.
\newblock Gaussian error linear units {(GELUs)}.
\newblock \emph{arXiv preprint arXiv:1606.08415}.

\bibitem[{Hendrycks and Gimpel(2017)}]{Hendrycks2016ABF}
Dan Hendrycks and Kevin Gimpel. 2017.
\newblock A baseline for detecting misclassified and out-of-distribution
  examples in neural networks.
\newblock In \emph{ICLR}.

\bibitem[{Hendrycks et~al.(2019{\natexlab{a}})Hendrycks, Lee, and
  Mazeika}]{hendrycks2019pretraining}
Dan Hendrycks, Kimin Lee, and Mantas Mazeika. 2019{\natexlab{a}}.
\newblock Using pre-training can improve model robustness and uncertainty.
\newblock \emph{ICML}.

\bibitem[{Hendrycks et~al.(2019{\natexlab{b}})Hendrycks, Mazeika, and
  Dietterich}]{Hendrycks2018DeepAD}
Dan Hendrycks, Mantas Mazeika, and Thomas~G. Dietterich. 2019{\natexlab{b}}.
\newblock Deep anomaly detection with outlier exposure.
\newblock \emph{ICLR}.

\bibitem[{Hendrycks et~al.(2019{\natexlab{c}})Hendrycks, Mazeika, Kadavath, and
  Song}]{hendrycks2019self}
Dan Hendrycks, Mantas Mazeika, Saurav Kadavath, and Dawn Song.
  2019{\natexlab{c}}.
\newblock Using self-supervised learning can improve model robustness and
  uncertainty.
\newblock In \emph{NeurIPS}.

\bibitem[{Hendrycks et~al.(2020)Hendrycks, Mu, Cubuk, Zoph, Gilmer, and
  Lakshminarayanan}]{hendrycks2020augmix}
Dan Hendrycks, Norman Mu, Ekin~D. Cubuk, Barret Zoph, Justin Gilmer, and Balaji
  Lakshminarayanan. 2020.
\newblock {AugMix}: A simple data processing method to improve robustness and
  uncertainty.
\newblock \emph{ICLR}.

\bibitem[{Hendrycks et~al.(2019{\natexlab{d}})Hendrycks, Zhao, Basart,
  Steinhardt, and Song}]{Hendrycks2019NaturalAE}
Dan Hendrycks, Kevin Zhao, Steven Basart, Jacob Steinhardt, and Dawn Song.
  2019{\natexlab{d}}.
\newblock Natural adversarial examples.
\newblock \emph{ArXiv}, abs/1907.07174.

\bibitem[{Hochreiter and Schmidhuber(1997)}]{hochreiter1997lstm}
Sepp Hochreiter and J{\"u}rgen Schmidhuber. 1997.
\newblock Long short-term memory.
\newblock In \emph{Neural Computation}.

\bibitem[{Jia and Liang(2017)}]{Jia2017AdversarialEF}
Robin Jia and Percy Liang. 2017.
\newblock Adversarial examples for evaluating reading comprehension systems.
\newblock In \emph{EMNLP}.

\bibitem[{Lan et~al.(2020)Lan, Chen, Goodman, Gimpel, Sharma, and
  Soricut}]{Lan2020ALBERT}
Zhenzhong Lan, Mingda Chen, Sebastian Goodman, Kevin Gimpel, Piyush Sharma, and
  Radu Soricut. 2020.
\newblock {ALBERT:} a lite {BERT} for self-supervised learning of language
  representations.
\newblock In \emph{ICLR}.

\bibitem[{Lang(1995)}]{Lang95}
Ken Lang. 1995.
\newblock {NewsWeeder}: Learning to filter {Netnews}.
\newblock In \emph{ICML}.

\bibitem[{Lee et~al.(2018)Lee, Lee, Lee, and Shin}]{kimin}
Kimin Lee, Honglak Lee, Kibok Lee, and Jinwoo Shin. 2018.
\newblock Training confidence-calibrated classifiers for detecting
  out-of-distribution samples.
\newblock In \emph{ICLR}.

\bibitem[{Li et~al.(2020)Li, Wallace, Shen, Lin, Keutzer, Klein, and
  Gonzalez}]{li2020train}
Zhuohan Li, Eric Wallace, Sheng Shen, Kevin Lin, Kurt Keutzer, Dan Klein, and
  Joseph~E Gonzalez. 2020.
\newblock Train large, then compress: Rethinking model size for efficient
  training and inference of transformers.
\newblock \emph{ArXiv}, abs/2002.11794.

\bibitem[{Liu et~al.(2019{\natexlab{a}})Liu, Schwartz, and
  Smith}]{liu2019inoculation}
Nelson~F Liu, Roy Schwartz, and Noah~A Smith. 2019{\natexlab{a}}.
\newblock Inoculation by fine-tuning: A method for analyzing challenge
  datasets.
\newblock In \emph{NAACL}.

\bibitem[{Liu et~al.(2019{\natexlab{b}})Liu, Ott, Goyal, Du, Joshi, Chen, Levy,
  Lewis, Zettlemoyer, and Stoyanov}]{liu2019roberta}
Yinhan Liu, Myle Ott, Naman Goyal, Jingfei Du, Mandar Joshi, Danqi Chen, Omer
  Levy, Mike Lewis, Luke Zettlemoyer, and Veselin Stoyanov. 2019{\natexlab{b}}.
\newblock {RoBERTa}: A robustly optimized {BERT} pretraining approach.
\newblock \emph{ArXiv}, abs/1907.11692.

\bibitem[{Maas et~al.(2011)Maas, Daly, Pham, Huang, Ng, and
  Potts}]{maas2011learning}
Andrew~L Maas, Raymond~E Daly, Peter~T Pham, Dan Huang, Andrew~Y Ng, and
  Christopher Potts. 2011.
\newblock Learning word vectors for sentiment analysis.
\newblock In \emph{ACL}.

\bibitem[{McAuley et~al.(2015)McAuley, Targett, Shi, and van~den
  Hengel}]{DBLP:journals/corr/McAuleyTSH15}
Julian~J. McAuley, Christopher Targett, Qinfeng Shi, and Anton van~den Hengel.
  2015.
\newblock Image-based recommendations on styles and substitutes.
\newblock In \emph{SIGIR}.

\bibitem[{Mikolov et~al.(2013)Mikolov, Sutskever, Chen, Corrado, and
  Dean}]{mikolov2013distributed}
Tomas Mikolov, Ilya Sutskever, Kai Chen, Greg~S Corrado, and Jeff Dean. 2013.
\newblock Distributed representations of words and phrases and their
  compositionality.
\newblock In \emph{NIPS}.

\bibitem[{Min et~al.(2019)Min, Wallace, Singh, Gardner, Hajishirzi, and
  Zettlemoyer}]{min2019compositional}
Sewon Min, Eric Wallace, Sameer Singh, Matt Gardner, Hannaneh Hajishirzi, and
  Luke Zettlemoyer. 2019.
\newblock Compositional questions do not necessitate multi-hop reasoning.
\newblock In \emph{ACL}.

\bibitem[{Naik et~al.(2018)Naik, Ravichander, Sadeh, Rose, and
  Neubig}]{naik2018stress}
Aakanksha Naik, Abhilasha Ravichander, Norman Sadeh, Carolyn Rose, and Graham
  Neubig. 2018.
\newblock Stress test evaluation for natural language inference.
\newblock In \emph{COLING}.

\bibitem[{Orhan(2019)}]{Orhan2019RobustnessPO}
A.~Emin Orhan. 2019.
\newblock Robustness properties of facebook's {ResNeXt WSL} models.
\newblock \emph{ArXiv}, abs/1907.07640.

\bibitem[{Pennington et~al.(2014)Pennington, Socher, and
  Manning}]{pennington2014glove}
Jeffrey Pennington, Richard Socher, and Christopher~D. Manning. 2014.
\newblock {GloVe}: Global vectors for word representation.
\newblock In \emph{EMNLP}.

\bibitem[{Quionero-Candela et~al.(2009)Quionero-Candela, Sugiyama,
  Schwaighofer, and Lawrence}]{QuioneroCandela2009DatasetSI}
Joaquin Quionero-Candela, Masashi Sugiyama, Anton Schwaighofer, and Neil~D.
  Lawrence. 2009.
\newblock Dataset shift in machine learning.

\bibitem[{Ribeiro et~al.(2018)Ribeiro, Singh, and
  Guestrin}]{ribeiro2018semantically}
Marco~Tulio Ribeiro, Sameer Singh, and Carlos Guestrin. 2018.
\newblock Semantically equivalent adversarial rules for debugging {NLP} models.
\newblock In \emph{ACL}.

\bibitem[{Sanh et~al.(2019)Sanh, Debut, Chaumond, and
  Wolf}]{sanh2019distilbert}
Victor Sanh, Lysandre Debut, Julien Chaumond, and Thomas Wolf. 2019.
\newblock {DistilBERT}, a distilled version of bert: smaller, faster, cheaper
  and lighter.
\newblock In \emph{NeurIPS EMC$^2$ Workshop}.

\bibitem[{Shen et~al.(2020)Shen, Dong, Ye, Ma, Yao, Gholami, Mahoney, and
  Keutzer}]{shen2019q}
Sheng Shen, Zhen Dong, Jiayu Ye, Linjian Ma, Zhewei Yao, Amir Gholami,
  Michael~W Mahoney, and Kurt Keutzer. 2020.
\newblock {Q-BERT}: Hessian based ultra low precision quantization of {BERT}.

\bibitem[{Socher et~al.(2013)Socher, Perelygin, Wu, Chuang, Manning, Ng, and
  Potts}]{socher2013recursive}
Richard Socher, Alex Perelygin, Jean Wu, Jason Chuang, Christopher~D Manning,
  Andrew Ng, and Christopher Potts. 2013.
\newblock Recursive deep models for semantic compositionality over a sentiment
  treebank.
\newblock In \emph{EMNLP}.

\bibitem[{Torralba and Efros(2011)}]{Torralba2011UnbiasedLA}
Antonio Torralba and Alexei~A. Efros. 2011.
\newblock Unbiased look at dataset bias.
\newblock \emph{CVPR}.

\bibitem[{Vaswani et~al.(2017)Vaswani, Shazeer, Parmar, Uszkoreit, Jones,
  Gomez, Kaiser, and Polosukhin}]{vaswani2017attention}
Ashish Vaswani, Noam Shazeer, Niki Parmar, Jakob Uszkoreit, Llion Jones,
  Aidan~N Gomez, {\L}ukasz Kaiser, and Illia Polosukhin. 2017.
\newblock Attention is all you need.
\newblock In \emph{NIPS}.

\bibitem[{Wallace et~al.(2019)Wallace, Feng, Kandpal, Gardner, and
  Singh}]{Wallace2019UniversalAT}
Eric Wallace, Shi Feng, Nikhil Kandpal, Matthew Gardner, and Sameer Singh.
  2019.
\newblock Universal adversarial triggers for attacking and analyzing {NLP}.
\newblock In \emph{EMNLP}.

\bibitem[{Wang et~al.(2019)Wang, Singh, Michael, Hill, Levy, and
  Bowman}]{Wang2018GLUEA}
Alex Wang, Amanpreet Singh, Julian Michael, Felix Hill, Omer Levy, and
  Samuel~R. Bowman. 2019.
\newblock {GLUE}: A multitask benchmark and analysis platform for natural
  language understanding.
\newblock In \emph{ICLR}.

\bibitem[{Wieting et~al.(2016)Wieting, Bansal, Gimpel, and
  Livescu}]{Wieting2015TowardsUP}
John Wieting, Mohit Bansal, Kevin Gimpel, and Karen Livescu. 2016.
\newblock Towards universal paraphrastic sentence embeddings.
\newblock In \emph{ICLR}.

\bibitem[{Williams et~al.(2018)Williams, Nangia, and
  Bowman}]{williams2017broad}
Adina Williams, Nikita Nangia, and Samuel~R Bowman. 2018.
\newblock A broad-coverage challenge corpus for sentence understanding through
  inference.
\newblock In \emph{NAACL-HLT}.

\bibitem[{Wolf et~al.(2019)Wolf, Debut, Sanh, Chaumond, Delangue, Moi, Cistac,
  Rault, Louf, Funtowicz, and Brew}]{Wolf2019HuggingFacesTS}
Thomas Wolf, Lysandre Debut, Victor Sanh, Julien Chaumond, Clement Delangue,
  Anthony Moi, Pierric Cistac, Tim Rault, R{\'e}mi Louf, Morgan Funtowicz, and
  Jamie Brew. 2019.
\newblock {HuggingFace's Transformers}: State-of-the-art natural language
  processing.
\newblock \emph{ArXiv}, abs/1910.03771.

\bibitem[{Xie and Yuille(2020)}]{Xie2019IntriguingPO}
Cihang Xie and Alan~L. Yuille. 2020.
\newblock Intriguing properties of adversarial training at scale.
\newblock In \emph{ICLR}.

\bibitem[{Xie et~al.(2020)Xie, Hovy, Luong, and Le}]{Xie2019SelftrainingWN}
Qizhe Xie, Eduard~H. Hovy, Minh-Thang Luong, and Quoc~V. Le. 2020.
\newblock Self-training with noisy student improves imagenet classification.
\newblock In \emph{CVPR}.

\bibitem[{Yogatama et~al.(2019)Yogatama, de~Masson~d'Autume, Connor,
  Kocisk{\'y}, Chrzanowski, Kong, Lazaridou, Ling, Yu, Dyer, and
  Blunsom}]{Yogatama2019LearningAE}
Dani Yogatama, Cyprien de~Masson~d'Autume, Jerome Connor, Tom{\'a}s
  Kocisk{\'y}, Mike Chrzanowski, Lingpeng Kong, Angeliki Lazaridou, Wang Ling,
  Lei Yu, Chris Dyer, and Phil Blunsom. 2019.
\newblock Learning and evaluating general linguistic intelligence.
\newblock \emph{ArXiv}, abs/1901.11373.

\bibitem[{Zhang et~al.(2018)Zhang, Liu, Liu, Gao, Duh, and
  Durme}]{DBLP:journals/corr/abs-1810-12885}
Sheng Zhang, Xiaodong Liu, Jingjing Liu, Jianfeng Gao, Kevin Duh, and
  Benjamin~Van Durme. 2018.
\newblock {ReCoRD:} bridging the gap between human and machine commonsense
  reading comprehension.
\newblock \emph{arXiv}, abs/1810.12885.

\end{thebibliography}
\bibliographystyle{acl_natbib}

\newpage
\appendix
\section{Additional Experimental Results}\label{appendix:additional}

\subsection{Significant OOD Accuracy Drops}

For STS-B, ReCoRD, and SST-2/IMDb, there is a noticeable drop in accuracy when testing on OOD examples. We show the STS-B results in Table~\ref{tab:STS-B}, the ReCoRD results in Table~\ref{tab:record}, and the SST-2/IMDb results in Table~\ref{tab:sst2}.

\subsection{Minor OOD Accuracy Drops}\label{appendix:minor}

\begin{figure}[h!]
    \centering
    \includegraphics[width=0.5\textwidth]{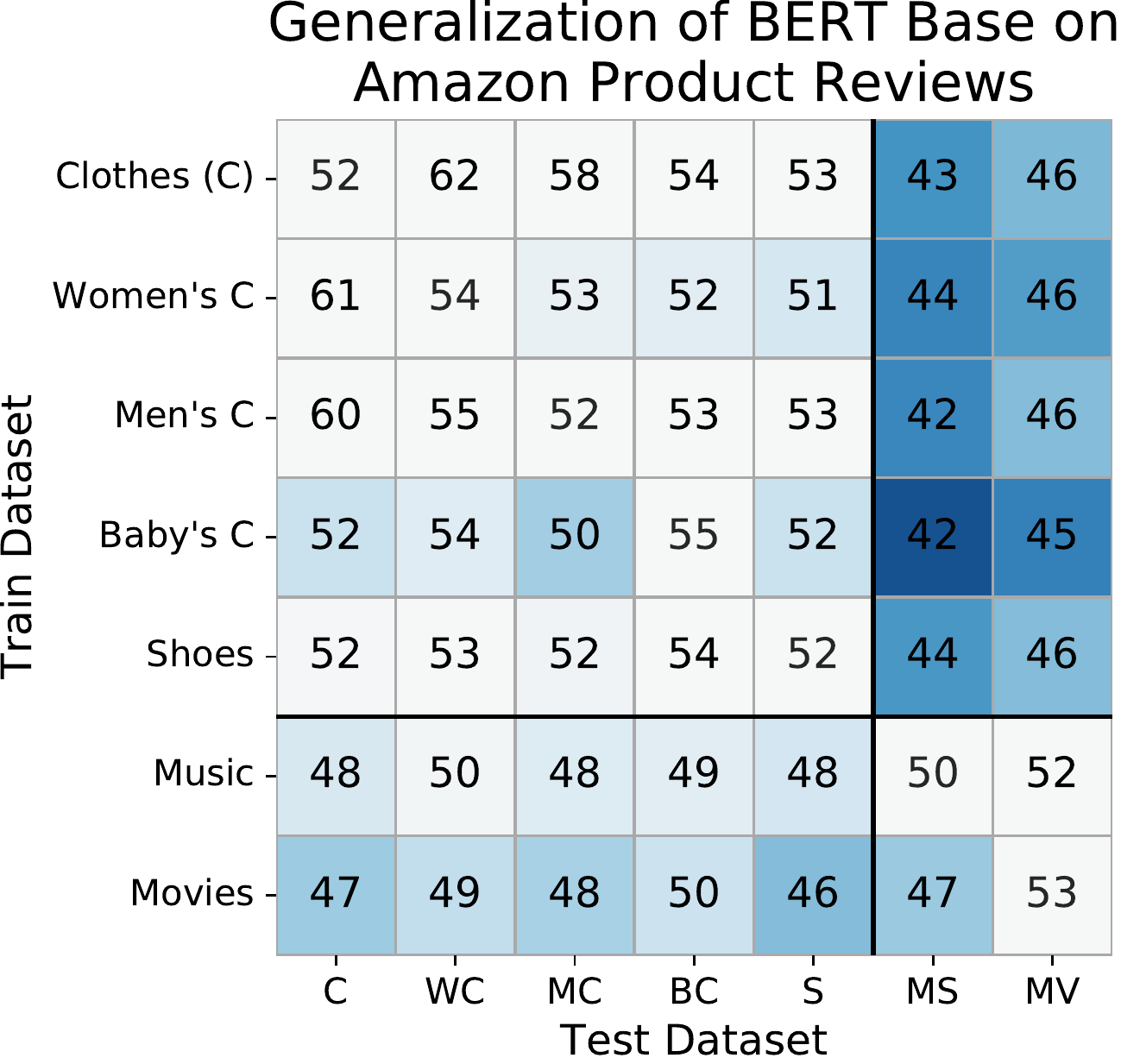}
    \caption{We finetune BERT Base on one category of Amazon reviews and then evaluate it on other categories. Models predict the review's star rating with 5-way classification. We use five clothing categories: Clothes (C), Women's Clothing (WC), Men's Clothing (MC), Baby Clothing (BC), and Shoes (S); and two entertainment categories: Music (MS), Movies (MV). BERT is robust for closely related categories such as men's, women's, and baby clothing. However, BERT struggles when there is an extreme distribution shift such as Baby Clothing to Music (dark blue region). Note this shift is closer to a domain adaptation setting.}
    \label{fig:amazon}
\end{figure}

We observe more minor performance declines for the Amazon, MNLI, and Yelp datasets. Figure~\ref{fig:amazon} shows the Amazon results for BERT Base, Table~\ref{tab:mnli} shows the MNLI results, and Table~\ref{tab:yelp} shows the Yelp results.

\subsection{OOD Detection}\label{appendix:ooddetection}
\begin{figure*}[h]
    \centering
    \includegraphics[width=\textwidth]{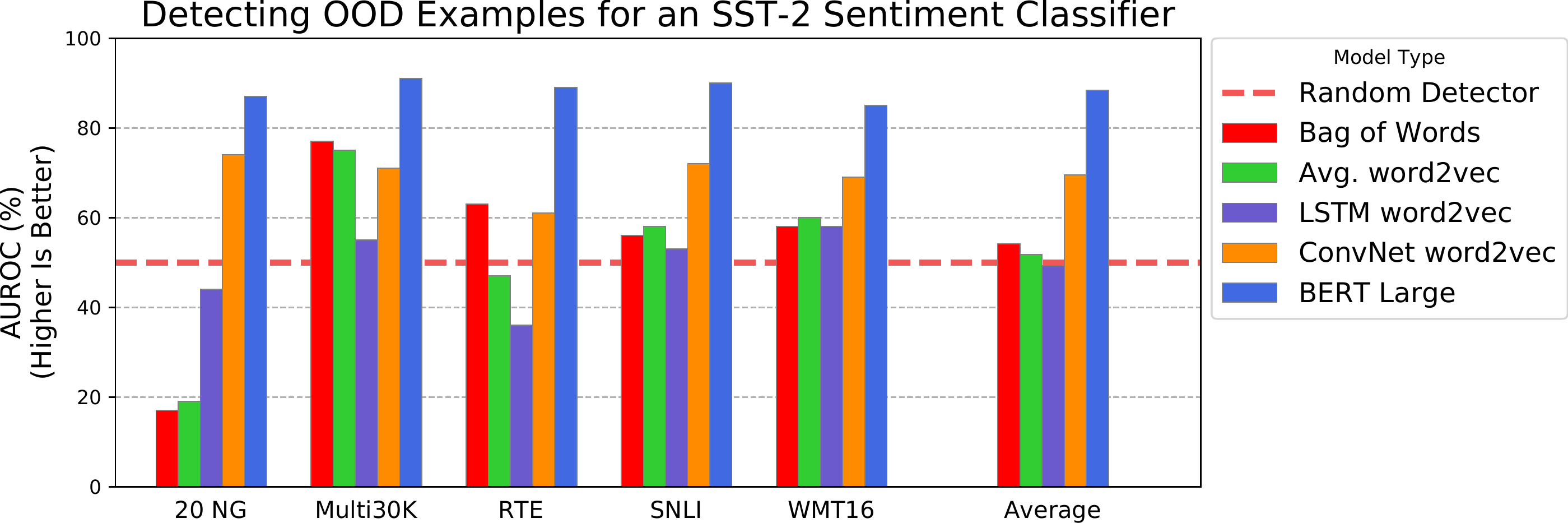}
    \caption{We feed in OOD examples from out-of-distribution datasets (20 Newsgroups, Multi30K, etc.) to SST-2 sentiment classifiers and report the AUROC detection performance. A 50\% AUROC is the random chance level.
    }
    \label{fig:auroc}
\end{figure*}

Full FAR95 values are in Table~\ref{tab:oodfpr}. We also report the Area Under the Receiver Operating Characteristic (AUROC)~\cite{Hendrycks2016ABF}. The AUROC is the probability that an OOD example receives a higher anomaly score than an in-distribution example, viz.,\[\mathrm{P}(-\max_y p(y\mid x_\text{out}) > -\max_y p(y\mid x_\text{in})).\] A flawless AUROC is $100\%$ while $50\%$ is random chance. These results are in Figure~\ref{fig:auroc} and Table~\ref{tab:oodauroc}.

\begin{table*}[h]
\scriptsize
\setlength\tabcolsep{3pt}
\centering
\begin{tabularx}{\textwidth}{*{1}{>{\hsize=0.3\hsize}X} *{1}{>{\hsize=0.9cm}X }
| *{10}{>{\hsize=0.4\hsize}L}}
Train & Test & BoW & Average word2vec & LSTM word2vec & ConvNet word2vec & Average GloVe & LSTM GloVe & ConvNet GloVe & BERT ~~Base & BERT Large & RoBERTa \\ 
\Xhline{2\arrayrulewidth} 

\parbox[t]{20mm}{\multirow{2}{*}{Images}}
& Images & {39.7} & {61.4} & {75.7} & {81.8} & {61.2} & {79.8} & {81.8} & {91.8} & {92.8} & {94.2} \\
& MSRvid & {4.4} (-35.4) & {11.3} (-50.1) & {38.3} (-37.4) & {62.0} (-19.8) & {6.1} (-55.2) & {43.1} (-36.7) & {57.8} (-24.1) & {89.5} (-2.3) & {90.5} (-2.3) & {94.3} (0.1) \\
\Xhline{2\arrayrulewidth} 

\parbox[t]{20mm}{\multirow{2}{*}{MSRvid}}
& MSRvid & {60.7} & {68.7} & {85.9} & {85.0} & {66.8} & {85.6} & {87.4} & {92.4} & {93.9} & {94.9} \\
& Images & {19.3} (-41.4) & {23.7} (-44.9) & {45.6} (-40.2) & {54.3} (-30.7) & {11.1} (-55.7) & {49.0} (-36.6) & {51.9} (-35.4) & {85.8} (-6.6) &  {86.8} (-7.1) & {90.4} (-4.6) \\
\Xhline{2\arrayrulewidth} 

\parbox[t]{20mm}{\multirow{2}{*}{Headlines}}
& Headlines & {26.8} & {58.9} & {66.2} & {67.4} & {53.4} & {69.9} & {69.6} & {87.0} & {88.3} & {91.3} \\
& MSRpar & {10.1} (-16.7) & {19.1} (-39.7) & {-1.9} (-68.1) & {9.8} (-57.6) & {25.9} (-27.5) & {25.4} (-44.5) & {10.9} (-58.7) & {69.9} (-17.1) & {63.6} (-24.7) & {75.5} (-15.8) \\
\Xhline{2\arrayrulewidth} 

\parbox[t]{20mm}{\multirow{2}{*}{MSRpar}}
& MSRpar & {47.0} & {27.0} & {46.7} & {49.8} & {50.9} & {46.7} & {46.2} & {78.8} & {81.6} & {86.8} \\
& Headlines & {-9.7} (-56.7) & {12.7} (-14.4) & {10.3} (-36.5) & {23.7} (-26.1) & {7.0} (-43.9) & {15.6} (-31.1) & {30.6} (-15.6) & {73.0} (-5.8) & {71.7} (-9.9) & {83.9} (-2.9) \\
\Xhline{3\arrayrulewidth}
\end{tabularx}
\caption{We train and test models on different STS-B distributions (Images, MSR videos, Headlines, and MSR paraphrase). The severe drop in the Pearson correlation coefficient shows the consequence of a distribution shift. Models such as Average GloVe lose nearly all performance when out-of-distribution. RoBERTa does especially well in comparison to other models.}
\label{tab:STS-B}
\end{table*}

\begin{table*}[h]
\small
\setlength\tabcolsep{3pt}
\centering
\begin{tabularx}{\textwidth}{*{1}{>{\hsize=0.2\hsize}X} *{1}{>{\hsize=1.4cm}X }
| *{5}{>{\hsize=0.4\hsize}L}}
Train & Test & Document QA & DistilBERT & BERT Base & BERT Large & RoBERTa\\ 
\Xhline{2\arrayrulewidth} 
 \parbox[t]{50mm}{\multirow{2}{*}{CNN}}
 & CNN &  39.0 & 45.0 & 53.2 & 67.2 & 71.5 \\
 & DailyMail & 29.7 (-9.3) & 34.8 (-10.2) & 46.7 (-6.6) & 59.8 (-7.4) & 72.2 (0.7) \\
\Xhline{2\arrayrulewidth} 
 \parbox[t]{50mm}{\multirow{2}{*}{DailyMail}}
 & DailyMail & 30.8 & 36.7 & 48.2 &	61.2 & 73.0\\
 & CNN & 36.9 (6.2) & 43.9 (7.2) & 51.8 (3.6) & 65.5 (4.3) & 73.0 (0.0) \\
\Xhline{3\arrayrulewidth}
\end{tabularx}
\caption{For ReCoRD, the exact match performance is closely tethered to the test dataset, which suggests a difference in the difficulty of the two test sets. This gap can be bridged by larger Transformer models pretrained on more data.}
\label{tab:record}
\end{table*}

\begin{table*}[t]
\scriptsize
\setlength\tabcolsep{3pt}
\centering
\begin{tabularx}{\textwidth}{*{1}{>{\hsize=0.3\hsize}X} *{1}{>{\hsize=0.7cm}X }
| *{14}{>{\hsize=0.4\hsize}L}}
Train & Test & BoW & Average word2vec & LSTM word2vec & ConvNet word2vec & Average GloVe & LSTM GloVe & ConvNet GloVe & {BERT ~~~Base} & BERT Large & RoBERTa\\ 
\Xhline{2\arrayrulewidth} 
 \parbox[t]{20mm}{\multirow{2}{*}{SST}}
 & SST  & 80.6 & 81.4 & 87.5 & 85.3 & 80.3 & 87.4 & 84.8 & 91.9 & 93.6 & 95.6 \\
 & IMDb & 73.9 (-6.8) & 76.4 (-5.0) & 78.0 (-9.5) & 81.0 (-4.4) & 74.5 (-5.8) & 82.1 (-5.3) & 81.0 (-3.8) & 87.5 (-4.4) & 88.3 (-5.3) & 92.8 (-2.8) \\
\Xhline{2\arrayrulewidth} 
 \parbox[t]{20mm}{\multirow{2}{*}{IMDb}}
 & IMDb  & {85.9} & {84.8} & {89.9} & {91.0} & {83.5} & {91.3} & {91.0} & {91.8} & {92.9} & {94.3} \\
 & SST  & 78.3 (-7.6) & 68.5 (-16.3) & 63.7 (-26.3) & 83.0 (-8.0) & 77.5 (-6.1) & 79.9 (-11.4) & 80.0 (-10.9) & 87.6 (-4.3) & 88.6 (-4.3) & 91.0 (-3.4) \\
\Xhline{3\arrayrulewidth}
\end{tabularx}
\caption{We train and test models on SST-2 and IMDB. Notice IID accuracy is not perfectly predictive of OOD accuracy, so increasing IID benchmark performance does not necessarily yield superior OOD generalization.}
\label{tab:sst2}
\end{table*}

\begin{table*}[h!]
\small
\setlength\tabcolsep{3pt}
\centering
\begin{tabularx}{\textwidth}{*{1}{>{\hsize=0.3\hsize}X} *{1}{>{\hsize=2.2cm}X }
| *{4}{>{\hsize=0.4\hsize}L}}
Train & Test & DistilBERT & BERT Base & BERT Large & RoBERTa\\ 
\Xhline{3\arrayrulewidth} 
 \parbox[t]{50mm}{\multirow{3}{*}{Telephone}}
 & Telephone & {77.5} & {81.4} & {84.0} & {89.6} \\
 & Letters & {75.6} (-1.9) & {82.3} (0.9) & {85.1} (1.0) & {90.0} (0.4) \\
 & Face-to-face & {76.0} (-1.4) & {80.8} (-0.7) & {83.2} (-0.8) & {89.4} (-0.2) \\
\Xhline{3\arrayrulewidth}
\end{tabularx}
\caption{We train models on the MNLI Telephone dataset and test on the Telephone, Letters, and Face-to-face datasets. The difference in accuracies are quite small (and sometimes even positive) for all four models. This demonstrates that pretrained Transformers can withstand various types of shifts in the data distribution.}
\label{tab:mnli}
\end{table*}

\begin{table*}[h!]
\scriptsize
\setlength\tabcolsep{3pt}
\centering
\begin{tabularx}{\textwidth}{*{1}{>{\hsize=0.2\hsize}X} *{1}{>{\hsize=0.6cm}X }
| *{14}{>{\hsize=0.4\hsize}L}}
Train & Test & BoW & Average word2vec & LSTM word2vec & ConvNet word2vec & Average GloVe & LSTM GloVe & ConvNet GloVe & DistilBERT & BERT Base & BERT Large & RoBERTa\\ 
\Xhline{2\arrayrulewidth} 
\parbox[t]{50mm}{\multirow{4}{*}{AM}}
& AM & 87.2 & 85.6 & 88.0 & 89.6 & 85.0 & 88.0 & 91.2 & 90.0 & 90.8 & 91.0 & 93.0 \\
& CH & 82.4 (-4.8) & 80.4 (-5.2) & 87.2 (-0.8) & 88.6 (-1.0) & 75.1 (-9.9) & 88.4 (0.4) & 89.6 (-1.6) & 91.8 (1.8) & 91.0 (0.2) & 90.6 (-0.4) & 90.8 (-2.2) \\
& IT & 81.8 (-5.4) & 82.6 (-3.0) & 86.4 (-1.6) & 89.4 (-0.2) & 82.0 (-3.0) & 89.2 (1.2) & 89.6 (-1.6) & 92.6 (2.6) & 91.6 (0.8) & 91.2 (0.2) & 91.8 (-1.2) \\
& JA & 84.2 (-3.0) & 86.0 (0.4) & 89.6 (1.6) & 89.4 (-0.2) & 79.2 (-5.8) & 87.8 (-0.2) & 89.2 (-2.0) & 92.0 (2.0) & 92.0 (1.2) & 92.2 (1.2) & 93.4 (0.4) \\
\Xhline{2\arrayrulewidth} 

\parbox[t]{50mm}{\multirow{4}{*}{CH}}
& CH & {82.2} & {84.4} & {87.6} & {88.8} & {84.4} & {89.2} & {89.0} & {90.2} & {90.4} & {90.8} & {92.4} \\
& AM & 82.2 (0.0) & 85.4 (1.0) & 88.0 (0.4) & 89.2 (0.4) & 83.0 (-1.4) & 85.6 (-3.6) & 90.2 (1.2) & 90.6 (0.4) & 88.8 (-1.6) & 91.8 (1.0) & 92.4 (0.0) \\
& IT & 84.6 (2.4) & 82.0 (-2.4) & 88.0 (0.4) & 89.6 (0.8) & 84.6 (0.2) & 88.6 (-0.6) & 90.4 (1.4) & 91.4 (1.2) & 89.0 (-1.4) & 90.2 (-0.6) & 92.6 (0.2) \\
& JA & 83.8 (1.6) & 85.8 (1.4) & 88.6 (1.0) & 89.0 (0.2) & 86.8 (2.4) & 88.8 (-0.4) & 89.6 (0.6) & 91.6 (1.4) & 89.4 (-1.0) & 91.6 (0.8) & 92.2 (-0.2) \\
\Xhline{2\arrayrulewidth} 

\parbox[t]{50mm}{\multirow{4}{*}{IT}}
& IT & {87.2} & {86.8} & {89.6} & {90.8} & {86.2} & {89.6} & {90.8} & {92.4} & {91.6} & {91.8} & {94.2} \\
& AM & 85.4 (-1.8) & 83.8 (-3.0) & 89.0 (-0.6) & 90.2 (-0.6) & 85.6 (-0.6) & 89.0 (-0.6) & 90.2 (-0.6) & 90.4 (-2.0) & 90.6 (-1.0) & 89.4 (-2.4) & 92.0 (-2.2) \\
& CH & 79.6 (-7.6) & 81.6 (-5.2) & 83.8 (-5.8) & 88.4 (-2.4) & 78.0 (-8.2) & 83.2 (-6.4) & 85.8 (-5.0) & 90.4 (-2.0) & 89.6 (-2.0) & 90.0 (-1.8) & 92.4 (-1.8) \\
& JA & 82.0 (-5.2) & {84.6} (-2.2) & {87.4} (-2.2) & {88.6} (-2.2) & {85.0} (-1.2) & {86.8} (-2.8) & {89.4} (-1.4) & {91.8} (-0.6) & {91.4} (-0.2) & {91.2} (-0.6) & {92.2} (-2.0) \\
\Xhline{2\arrayrulewidth} 

\parbox[t]{50mm}{\multirow{4}{*}{JA}}
& JA & {85.0} & {87.6} & {89.0} & {90.4} & {88.0} & {89.0} & {89.6} & {91.6} & {92.2} & {93.4} & {92.6} \\
& AM & {83.4} (-1.6) & {85.0} (-2.6) & {87.8} (-1.2) & {87.8} (-2.6) & {80.4} (-7.6) & {88.6} (-0.4) & {89.4} (-0.2) & {91.2} (-0.4) & {90.4} (-1.8) & {90.6} (-2.8) & {91.0} (-1.6) \\
& CH & {81.6} (-3.4) & {83.6} (-4.0) & {89.0} (0.0) & {89.0} (-1.4) & {80.6} (-7.4) & {87.4} (-1.6) & {89.2} (-0.4) & {92.8} (1.2) & {91.4} (-0.8) & {90.8} (-2.6) & {92.4} (-0.2) \\
& IT & {84.0} (-1.0) & {83.6} (-4.0) & {88.2} (-0.8) & {89.4} (-1.0) & {83.6} (-4.4) & {88.0} (-1.0) & {90.6} (1.0) & {92.6} (1.0) & {90.2} (-2.0) & {91.0} (-2.4) & {92.6} (0.0) \\
\Xhline{3\arrayrulewidth}
\end{tabularx}
\caption{We train and test models on American (AM), Chinese (CH), Italian (IT), and Japanese (JA) restaurant reviews. The accuracy drop is smaller compared to SST-2/IMDb for most models and pretrained transformers are typically the most robust.}
\label{tab:yelp}
\end{table*}

\begin{table*}
\vspace{-5pt}
\small
\setlength\tabcolsep{1.5pt}
\centering
\begin{tabularx}{\textwidth}{*{1}{>{\hsize=0.25\hsize}X} *{1}{>{\hsize=1.4cm}X }
| *{11}{>{\hsize=0.4\hsize}Y}}
$\mathcal{D}_\text{in}$ & \multicolumn{1}{l|}{$\mathcal{D}_\text{out}^\text{test}$} &

\multicolumn{1}{p{1cm}}{\centering BoW} &
\multicolumn{1}{p{1cm}}{\centering Avg w2v} & \multicolumn{1}{p{1cm}}{\centering Avg GloVe} &
\multicolumn{1}{p{1cm}}{\centering LSTM w2v} & \multicolumn{1}{p{1cm}}{\centering LSTM GloVe} & 
\multicolumn{1}{p{1cm}}{\centering ConvNet w2v} & \multicolumn{1}{p{1cm}}{\centering ConvNet GloVe}  &
\multicolumn{1}{p{1.35cm}}{\centering DistilBERT} & \multicolumn{1}{p{1cm}}{\centering BERT Base} &
\multicolumn{1}{p{1cm}}{\centering BERT Large} & \multicolumn{1}{p{1cm}}{\centering RoBERTa}
\\ \hline

\parbox[t]{50mm}{\multirow{5}{*}{SST}}
 & 20 NG     & 100 & 100 & 100 & 94 & 90 & 61 & 71 & 39 & 35 & 29 & 22 \\
 & Multi30K  & 61  & 57  & 52  & 92 & 85 & 65 & 63 & 37 & 22 & 23 & 61 \\
 & RTE       & 100 & 100 & 84  & 93 & 88 & 75 & 56 & 43 & 32 & 29 & 36 \\
 & SNLI      & 81  & 83  & 72  & 92 & 82 & 63 & 63 & 38 & 28 & 28 & 29 \\
 & WMT16     & 100 & 91  & 77  & 90 & 82 & 70 & 63 & 56 & 48 & 44 & 65 \\
\Xhline{0.5\arrayrulewidth} \multicolumn{2}{c|}{Mean FAR95} & {88.4} & {86.2} & {76.9} & {92.2} & {85.4} & {66.9} & {63.1} & {42.5} & {33.0} & {\textbf{30.5}} & {43.0} \\
\Xhline{3\arrayrulewidth}
\end{tabularx}
\caption{Out-of-distribution detection FAR95 scores for various NLP models using the maximum softmax probability anomaly score. Observe that while pretrained Transformers are consistently best, there remains room for improvement.}
\label{tab:oodfpr}
\end{table*}

\begin{table*}
\vspace{-5pt}
\small
\setlength\tabcolsep{1.5pt}
\centering
\begin{tabularx}{\textwidth}{*{1}{>{\hsize=0.25\hsize}X} *{1}{>{\hsize=1.4cm}X }
| *{11}{>{\hsize=0.4\hsize}Y}}
$\mathcal{D}_\text{in}$ & \multicolumn{1}{l|}{$\mathcal{D}_\text{out}^\text{test}$} &

\multicolumn{1}{p{1cm}}{\centering BoW} &
\multicolumn{1}{p{1cm}}{\centering Avg w2v} & \multicolumn{1}{p{1cm}}{\centering Avg GloVe} &
\multicolumn{1}{p{1cm}}{\centering LSTM w2v} & \multicolumn{1}{p{1cm}}{\centering LSTM GloVe} & 
\multicolumn{1}{p{1cm}}{\centering ConvNet w2v} & \multicolumn{1}{p{1cm}}{\centering ConvNet GloVe}  &
\multicolumn{1}{p{1.35cm}}{\centering DistilBERT} & \multicolumn{1}{p{1cm}}{\centering BERT Base} &
\multicolumn{1}{p{1cm}}{\centering BERT Large} & \multicolumn{1}{p{1cm}}{\centering RoBERTa}
\\ \hline

\parbox[t]{50mm}{\multirow{5}{*}{SST}}
 & 20 NG & 17 & 19 & 30 & 44 & 59 & 74 & 64 & 82 & 83 & 87 & 90 \\
 & Multi30K  & 77 & 75 & 80 & 55 & 62 & 71 & 73 & 86 & 93 & 91 & 89 \\
 & RTE     & 63 & 47 & 72 & 36 & 54 & 61 & 77 & 83 & 89 & 89 & 90 \\
 & SNLI     & 56 & 58 & 71 & 53 & 64 & 72 & 74 & 86 & 92 & 90 & 92 \\
 & WMT16  & 58 & 60 & 69 & 58 & 63 & 69 & 74 & 80 & 85 & 85 & 83 \\
\Xhline{0.5\arrayrulewidth} \multicolumn{2}{c|}{Mean AUROC} & {54.2} & {51.8} & {64.5} & {49.3} & {60.4} & {69.5} & {72.5} & {83.1} & {88.1} & {88.4} & {\textbf{88.7}} \\
\Xhline{3\arrayrulewidth}
\end{tabularx}
\caption{Out-of-distribution detection AUROC scores for various NLP models using the maximum softmax probability anomaly score. An AUROC score of $50\%$ is random chance, while $100\%$ is perfect.}
\label{tab:oodauroc}
\end{table*}

\end{document}